# Finite State Machine Synthesis for Evolutionary Hardware

Andrey Bereza, Maksim Lyashov, Luis Blanco
*Dept. of Information Systems and Radio Engineering, Don State Technical University*
*anbirch@mail.ru, maxl85@mail.ru, raubtierxxx@gmail.com*

## Abstract

*This article considers application of genetic algorithms for finite machine synthesis. The resulting genetic finite state machines synthesis algorithm allows for creation of machines with less number of states and within shorter time. This makes it possible to use hardware-oriented genetic finite machines synthesis algorithm in autonomous systems on reconfigurable platforms.*

## 1. Introduction

Nowadays evolutionary algorithms (EA) are applied to design digital and analog devices [1]. This trend is called evolutionary electronics [2, 3]. The application of EA on hardware platforms with reconfigurable elements, which allows for rebuilding the systems in process of operation, is called evolutionary hardware [4]. Evolutionary hardware (EH) is a new type of hardware based on various probabilistic algorithms such as genetic algorithms (GA) and evolutionary programming.

In EH design the reconfigurable parts are dynamically rebuilding combinatory or sequential logic circuits [4]. To dynamically rebuild digital logic circuit it is necessary for GA to be able to synthesize circuit on a gate level. Hence, the task of digital logic circuit synthesis EA development arises.

Current methods of finite state machine synthesis always use the specifics of a problem, which makes it impossible to use that same state machine generation technic for any different kind of problem. The quest is to make universal state machine synthesis method applicable to a wide range of problems. The application of EA for finite machine synthesis is shown in the work [5]. However given algorithms are applied for state machine programming, where program is described with finite state machines, which doesn't allow their usage in autonomous hardware systems or reconfigurable platforms.

* The research is supported by Russian Foundation for Basic Research (grant #13-07-00951)

## 2. The problem of state machine evolutionary synthesis

The problem of state machine evolutionary synthesis is defined as set:
$$R = \{H, O, F\},$$
where $H$ – the synthesized solution genotype, $O$ – the genetic operators $O = \{o_1, o_2, \ldots o_n\}$, $F$ – the objective function.

The synthesized solution genotype is defined as set:
$$H = \{g_1, g_2, \ldots g_p\},$$
where $p = S \cdot 2^x$, $S$ – the amount of finite machine states, $x$ – the amount of inputs.

The objective function is defined as expression:
$$F = w_1 a_1 + w_2 a_2,$$
where $a_1$ – the amount of states, $a_2$ – the amount of iterations, $w$ – the weight coefficients for particular criteria.

The task for GA is to minimize the objective function, e.g. $F \to min$.

## 3. Hardware-oriented genetic algorithm of finite state machine synthesis

Schematic diagram of proposed hardware oriented genetic algorithm of finite state machine synthesis, designed for EH creation on a reconfigurable platform is shown on a figure 1.

On the first step user sets requirements for finite state machine being designed. Those are the numbers of states and triggers. Also the amount of generations and mutation and crossing over probabilities must be set to organize the process of evolutionary synthesis. Then, according to the algorithm, initial set of solutions is generated and evaluated; transition correction algorithm is also executed. If during the evaluation process of initial set of solutions population has a solution that meets all the requirements, then it gets saved and algorithm ends. The terminal criteria of GA are reaching the maximum number of generations or having a solution.

Primary genetic operators in FSMGA are mutation and crossing over. Chromosome coding with bit string applies some constrains on operator types.

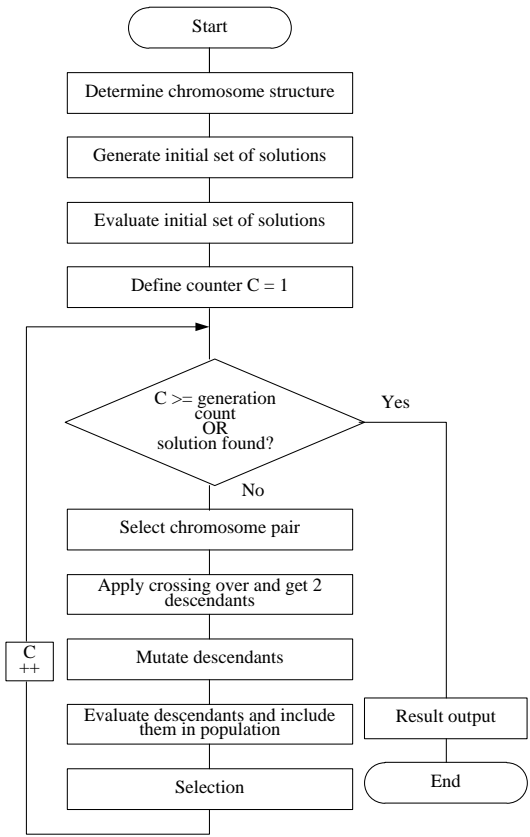

Figure 1 – Schematic diagram of hardware oriented genetic algorithm of finite state machine synthesis

Mutation operator is random, e.g. it does not depend on chromosome fitness or gene residing in chromosome. As the result of mutation it randomly changes either the output value of state machine or state number which will be selected by randomly picked transition.

Crossing over operator randomly exchanges genetic information between two solutions, while existing genetic information is being preserved. The GA solution quality is largely dependent on crossing over operator type selection. In proposed finite state machine synthesis algorithm one-point and two-point crossing over operators were applied, as those have the simplest hardware implementation [4]. Experimental studies have shown that two-point crossing over is preferable.

Selection operation algorithm is based on bubble sort algorithm, since its hardware implementation takes the least resources among other sorting algorithms [4]. After the population has been sorted in descending order (e.g. chromosomes with a higher objective function value are moved to the top of the population), 3 chromosomes with worst objective function value are deleted from population (since after crossing over and mutation operators 3 more chromosomes are inserted into population).

Since proposed GA is designed to function on autonomous EH, chromosomes are coded with bit strings. Let's consider chromosome coding on a specific example of finite machine.

Combinatory logic built on logical elements is replaced by RAM on the schematic diagram (figure 2). To do so, finite machine transition table should be converted first to be able to replace combinatory circuit with RAM. Transition graph of the finite machine is given on figure 3; it describes the behavior of some control device. The amount of triggers required to represent four states is equal to 2.

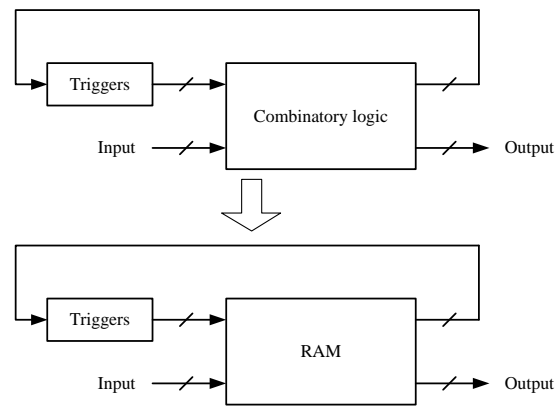

Figure 2 – Memory usage in state machine

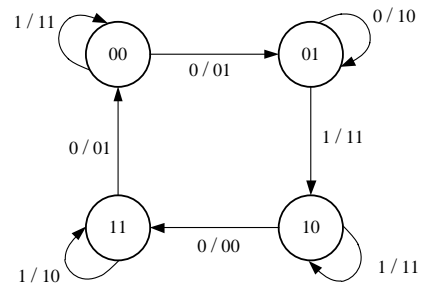

Figure 3 – State machine sample

For a state machine the number of RAM address inputs is equal to a sum of the amount of triggers and the number of inputs. RAM output number is accordingly equal to a sum of the amount of triggers and the number of outputs. For this finite machine sample RAM should contain 3 address inputs and 4 outputs; e.g. the required memory size is $8 \times 4$ bits. The truth table of the RAM is equal to that one of a state machine.

The schematic diagram for the state machine on a figure 3, which has been made by replacement of combinatory logics with RAM, is shown on figure 4.

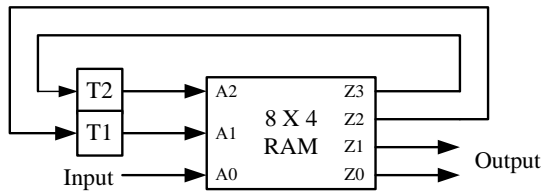

Figure 4 – Schematic diagram of state machine

## 4. Experimental studies

The developed algorithm has been tested on two different problems: «Santa Fe Trail» problem («Smart Ant») and autopilot construction for simplified helicopter model problem.

«Santa Fe Trail» – is a problem from the area of cooperative usage of GA and finite state machines [5]. The ant is on the surface of torus, which has size of 32x32 cells. The food is placed in some of the cells (on figure 5 marked as black). It is located along the broken line, but not in all cells. Broken line cells with no food are marked as gray. White cells do not belong to a broken line and contain no food. Altogether the field contains 89 food cells.

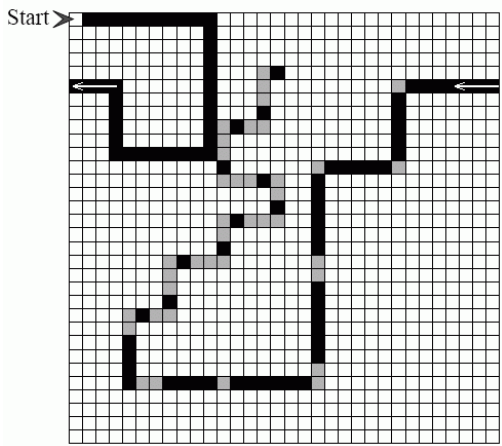

Figure 5 – The «Santa Fe Trail» field

Ant's starting location is marked with "Start". Ant occupies one cell and looks in one of four directions (up, down, left, right).

Ant is able to determine if the food is directly in front of him. In one game turn ant is able to make one of three actions:
− step forward, eating any food in destination;
− turn left;
− turn right.

The food eaten by the ant won't refill, the ant is always alive, the food isn't vital for him. Broken line isn't random, but strictly fixed. Ant is able to walk through any cell of field.

The game is 200 moves long; each move ant performs one of the three actions. After 200 turns the amount of eaten food is calculated. That is the result of the game.

The goal is to design an ant which will eat as much as possible food within 200 turns (all 89 is desirable).

One of the ways to describe the behavior of the ant is Mealy machine, which has one input variable (tells if the food is in front of the ant), and a set of output actions consisting of three, described above. Schematic diagram of this machine is given on figure 6.

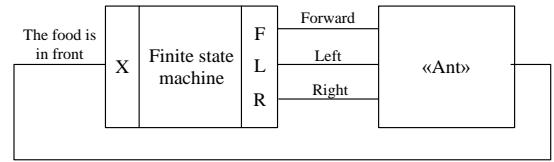

Figure 6 – Schematic diagram of «Smart Ant» finite state machine

It is hard to heuristically build a machine solving problem. For instance, a heuristically built Mealy machine with five states [6] can't solve this problem. Finite state machine describes the ant which eats only 81 food cells within 200 moves, and it takes 314 moves to eat all the food.

Experimental studies of «Santa Fe Trail» problem were conducted with population size of 1200, crossing over probability of 0.4 and mutation probability of 0.25. The comparison of the results of proposed finite machine synthesis GA (FMSGA) versus the results of heuristic algorithm and GA proposed in work [6] are shown in table 1.

Table 1 – Experimental results

| Algorithm | Amount of states | Moves | Synthesis time |
|---|---|---|---|
| Heuristic | 5 | 314 | - |
| GA[6] | 7 | 198 | 269 s. |
| FMSGA | 7 | 190 | *29 s.* |

As implied by the above results, FMSGA has been able to find a machine with 7 states which solves the problem in 190 moves. It also takes 9 times less amount of time to synthesize the machine then existing analogs do.

Consider the second problem of autopilot construction for simplified helicopter model. An autopilot has to be created for a simplified helicopter model which moves on a flat surface [5]. In one move helicopter model can either rotate through a certain predefined angle or change velocity.

The autopilot's task is to drive a helicopter through N markers within a limited time. The best autopilot is the one who manages to visit the highest number of markers. If two autopilots reach the same amount of markers, the one closest to a next marker at the last moment of the flight wins.

The autopilot input variable receives the sight sector number (figure 7). Current target position relative to a helicopter is given as an angle between helicopter's movement direction and the direction to a next marker (figure 7a). Helicopter always flies in the middle of the current sector. All sectors are static relative to a helicopter (figure 7b).

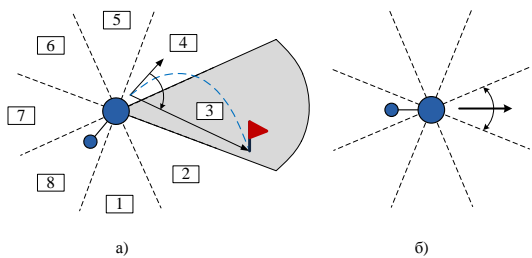

Figure 7 – Helicopter output data

Autopilot model is a finite machine with discrete input and output actions. Machine state indirectly maps helicopter's current position, its speed and history of state transitions.

Experimental studies have been conducted with sector sizes of 4 and 6. For each parameter set 50 tests have been conducted.

Experimental results are shown in table 2. The «Result» column shows amount of markers visited by autopilot designed with FMSGA. In work [5] the finite machine with 12 states is able to drive the helicopter through the first 18 out of 20 markers within given time.

Table 2 – Experimental results of helicopter autopilot design with FMSGA

| Number of sectors | Result | | |
|---|---|---|---|
| | Worst | Average | Best |
| 4 | 11 | 18 | 20 |
| 6 | 12 | 17 | 20 |

## 5. Conclusion

As shown by the given experimental results, developed FMSGA allows for machine synthesis within shorter time and with less number of states. This proves that developed hardware-oriented FMSGA can be effectively used in autonomous systems on reconfigurable platforms.